\documentclass[runningheads]{llncs}
\usepackage{graphicx}
\usepackage{times} 
\usepackage{listings}
\usepackage{xspace}
\usepackage{todonotes}
\usepackage[shortlabels]{enumitem}

\newcommand{\ie}{\textit{i}.\textit{e}.,\xspace}

\begin{document}
\title{NeSy4VRD: A Multifaceted Resource for Neurosymbolic AI Research using Knowledge Graphs in Visual Relationship Detection}
\titlerunning{NeSy4VRD: A Resource for NeSy Research using Knowledge Graphs}
%
\author{David Herron\inst{1}\orcidID{0009-0008-2736-6789} \and 
Ernesto Jim\'{e}nez-Ruiz\inst{1,2}\orcidID{0000-0002-9083-4599} \and  
Giacomo Tarroni\inst{1,3}\orcidID{0000-0002-0341-6138} \and  
Tillman Weyde\inst{1}\orcidID{0000-0001-8028-9905}}
\authorrunning{D. Herron et al.}
%
\institute{City, University of London, United Kingdom \\ 
\email{\{david.herron, ernesto.jimenez-ruiz, giacomo.tarroni, \\ t.e.weyde\}@city.ac.uk} \and
SIRIUS, University of Oslo, Norway \and 
Imperial College London, United Kingdom}
\maketitle    
\begin{abstract}
NeSy4VRD is a multifaceted resource designed to support the development of neurosymbolic AI (NeSy) research. 
NeSy4VRD re-establishes public access to the images of the VRD dataset and couples them with an extensively revised, quality-improved version of the VRD visual relationship annotations. 
Crucially, NeSy4VRD provides a well-aligned, companion OWL ontology that describes the dataset domain.
It comes with open source infrastructure that provides comprehensive support for extensibility of the annotations (which, in turn, facilitates extensibility of the ontology), and open source code for loading the annotations to/from a knowledge graph.
We are contributing NeSy4VRD to the computer vision, NeSy and Semantic Web communities to help foster more NeSy research using OWL-based knowledge graphs.


\end{abstract}
{
\small
\textbf{Resource type}: Image Dataset with Ontology and Infrastructure for Extensibility \\ 
\textbf{Licenses}: CC BY 4.0 (Dataset/Ontology), MIT/Expat (Infrastructure) \\
\textbf{Dataset and ontology}: \url{https://doi.org/10.5281/zenodo.7916355} \\
\textbf{Infrastructure}: \url{https://github.com/djherron/NeSy4VRD}
}
\section{Introduction}

Neurosymbolic AI (NeSy) explores combining neural learning with symbolic background knowledge and reasoning~\cite{Besold2017NeuralSymbolicLA/iswc2023}. 
Interest in NeSy is accelerating~\cite{DBLP:journals/corr/abs-2012-05876,hitzlerKGinNSI2021/splncs}.
The symbolic background knowledge representation and deductive reasoning capabilities of OWL ontologies and knowledge graphs (KGs) mark them as compelling candidates for participating as symbolic components in hybrid NeSy systems.

The Semantic Web has a strong association with the symbolic tradition of AI~\cite{DBLP:journals/cacm/Hitzler21}.
OWL-based knowledge graphs are exemplars of the explicit symbolic knowledge and symbol manipulation and reasoning machinery that Marcus~\cite{DBLP:journals/corr/abs-2002-06177} advocates be included in hybrid NeSy systems.
The many synergies between Semantic Web technologies and deep learning promise improved predictive performance~\cite{DBLP:journals/semweb/HitzlerBES20}.
Additionally, there are many benefits of OWL-based KGs for NeSy systems~\cite{Herron/NeSy2023}.

However, the amount of NeSy research using OWL-based KGs, particularly leveraging their reasoning capabilities, is low. 
A recent study of approaches to combining Semantic Web technologies with machine learning~\cite{Breit-and-Waltersdorfer}, involving nearly 500 papers, finds that only 29 of the papers report using `deductive components' of any kind, and of these only 20 report actually using `reasoning capabilities'.
Thus, the potential for OWL-based KGs to contribute to NeSy research is under-explored and, 
as we argue in~\cite{Herron/NeSy2023}, they merit a larger more prominent role in NeSy research.

An important factor in  explaining the low volume of NeSy research using OWL-based KGs, in addition to the cross-disciplinary nature of the endeavour, 
is surely that it has highly specialised prerequisites: a dataset for neural learning, plus a companion OWL ontology describing the domain of the data over which a knowledge graph enables pertinent reasoning.
Such specialised resources are scarce, especially in combination, and, for many application areas, may not exist.
The scarcity of appropriate dataset/ontology resources is hindering promising NeSy research that could use OWL-based KGs.

In this paper we present NeSy4VRD, a multifaceted resource designed to foster NeSy research using OWL-based KGs within the application area of visual relationship detection in images.
The paramount purpose of NeSy4VRD is to lower the barriers to entry for conducting NeSy research using OWL-based KGs.
NeSy4VRD builds on top of the Visual Relationship Detection (VRD) dataset \cite{lu2016visual} and provides an extensively revised, quality-improved version of its visual relationship annotations.
The main components and contributions of NeSy4VRD are as follows:
\begin{enumerate}[(i)]
    \item an image dataset with high-quality visual relationship annotations that reference a large number of object classes and predicates (relations);
    \item a well-aligned, companion OWL ontology, called VRD-World, that describes the domain of the images and visual relationships;
    \item sample Python code for loading the annotated visual relationships into a knowledge graph hosting the VRD-World ontology, and for extracting them from a knowledge graph and restoring them to their native format;
    \item support for extensibility of the annotations (and, thereby, the ontology) in the form of \textit{(a)} comprehensive Python code enabling deep but easy analysis of the images and their annotations, \textit{(b)} a custom, text-based protocol for specifying annotation customisation instructions declaratively, and \textit{(c)} a configurable, managed Python workflow for customising annotations in an automated, repeatable process.
\end{enumerate}

The remainder of this paper is structured as follows.     Section 2 outlines the background motivation for NeSy4VRD. Sections 3-6 describe
the four components of \linebreak NeSy4VRD mentioned above. Section 7 highlights intended and potential NeSy4VRD user groups and use cases. Finally, Section 8 concludes with summary remarks.

\section{Background to NeSy4VRD}

Our ambition was to conduct vision-based NeSy research using OWL-based KGs, but we were unable to find an image dataset with a companion OWL ontology appropriate for enabling our research vision.
During our search for an appropriate resource, the application task of visual relationship detection caught our attention. 
A number of image datasets exist that are accompanied by annotations (in various formats) describing visual relationships between ordered pairs of objects in the images, such as \cite{DBLP:journals/corr/KrishnaZGJHKCKL16,DBLP:journals/ijcv/KuznetsovaRAUKP20,lu2016visual,BMVC2015_52} and various dataset versions derived from these (as mentioned in~\cite{DBLP:journals/corr/abs-2201-00443}). 
But none of these has a companion OWL ontology describing the domain of the~images.

\subsection{The VRD dataset}

\begin{figure}
\includegraphics[width=0.98\textwidth]{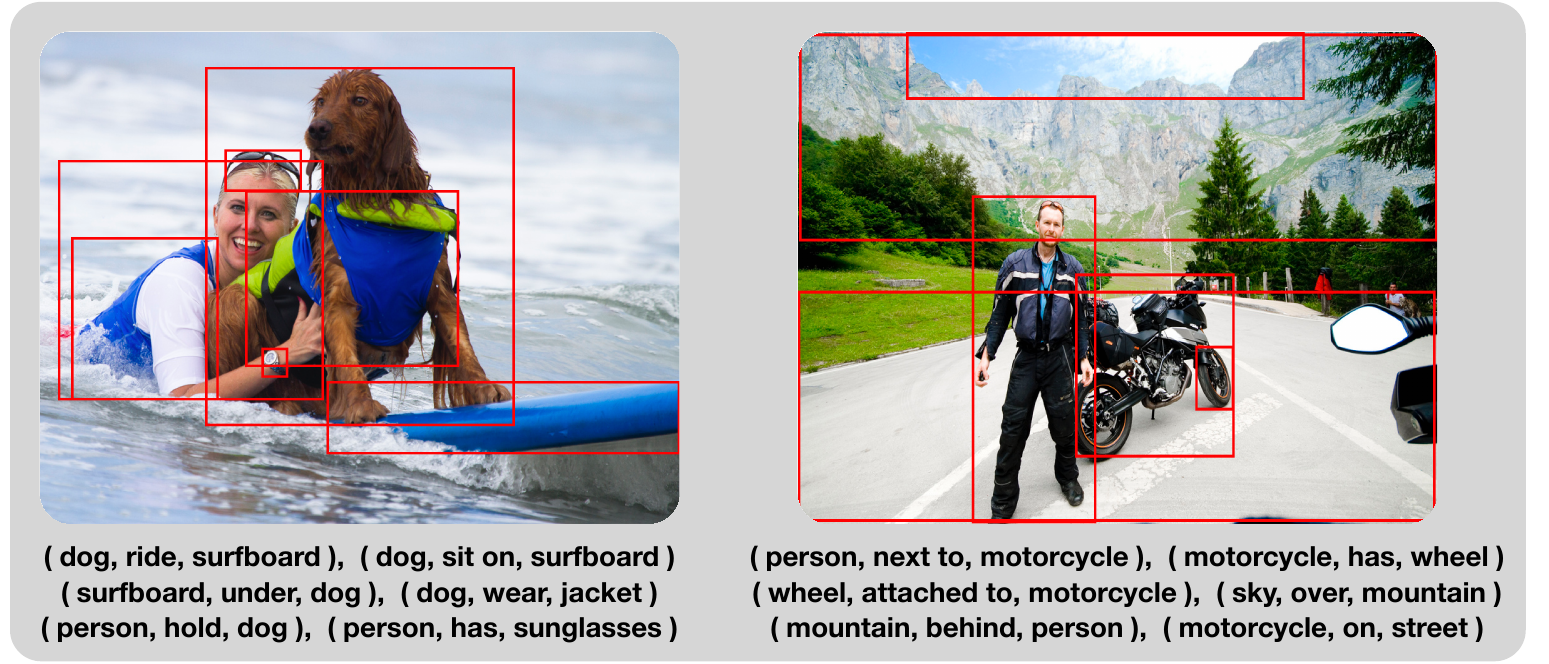}
\caption{Two representative images from the VRD dataset with samples of representative annotated visual relationships as 3-tuples.}
\label{fig:two-vrd-images}
\end{figure}

To advance our research, we selected the Visual Relationship Detection (VRD) image dataset \cite{lu2016visual} and resolved to engineer our own custom, companion OWL ontology. 
The visual relationships annotated for the VRD images are 5-tuples
\begin{verbatim}
 (subj_bbox, subj_class, predicate, obj_bbox, obj_class)
\end{verbatim}
that, for simplicity, can be thought of as 3-tuples, where the \texttt{subject} and \texttt{object} of each relationship are understood to be specified in terms of both a box (\ie \texttt{subj\_bbox} and \texttt{obj\_bbox)} and class (\ie \texttt{subj\_class} and \texttt{obj\_class}). 
The \texttt{predicate} describes some relation between the ordered pair of objects.
Figure~\ref{fig:two-vrd-images} shows two example VRD images with their objects and some of their annotated visual relationships.

\subsection{Attractive characteristics of the VRD dataset}
\label{sec:vrd-attractive}

Several characteristics make the VRD image dataset attractive, especially for NeSy research. 
The first is its small size: 4,000 training images and 1,000 test images. 
Since deep learning is known to be data hungry, the small amount of training data (in theory) leaves greater space for symbolic components (like OWL-based KGs) to enhancing deep learning's predictive performance. 
Second, relative to its size, the number of distinct (common) object classes and predicates (mostly common spatial relations, verbs and prepositions) referenced in its visual relationships is relatively large: 100 and 70, respectively. 
Third, the distribution of the \textit{types} of annotated visual relationships, 
\begin{equation}
    (s_i, p_k, o_j) \quad i,j = 1,\dots,100 \quad k = 1,\dots,70,
\end{equation}
has a long tail that provides plentiful zero/few-shot learning scenarios.
These scenarios represent ideal opportunities for NeSy researchers to explore ways by which symbolic components (like OWL-based KGs) can improve deep learning's ability to generalise.

The survey \cite{DBLP:journals/corr/abs-2201-00443} provides evidence that many researchers have been attracted to the VRD dataset. 
It also conveys the fact that the VRD dataset has supported research that has focused not only on the task of visual relationship detection in isolation but also on the broader, more general task of scene graph generation, which relies on visual relationship detection as a building block capability.
As explained in \cite{DBLP:journals/internet/KhanBC22,DBLP:journals/corr/abs-2201-00443}, scene graphs are commonly used as precursors to a variety of enriched, downstream visual understanding and reasoning application tasks, such as image captioning, visual question answering, image retrieval, image generation and multimedia event processing.

\subsection{Unattractive characteristics of the VRD dataset}
\label{sec:vrd-problems}

Motivated by a determination to engineer a companion OWL ontology for the VRD dataset that we knew to be robust, we embarked upon a deep analysis of the VRD features.
Despite its attractive characteristics and popularity, our analysis revealed that its visual relationship annotations contain many serious shortcomings.
Our primary concern was to ensure that we fully understood the semantics of the 100 object class names and 70 predicate names so that we could make reliable, informed decisions \textit{(a)} when constructing an OWL class hierarchy and \textit{(b)} when specifying object property characteristics and relationships, and when specifying classes for property domain and range restrictions.
For example, the object class name \texttt{plate} suggests dishware, but does it really mean that, and is that all it means?
Similarly, the predicate \texttt{across} feels familiar at first glance, but how is it really used, and is it used in one way consistently?
Unfortunately, the more our analysis progressed, the more the quantity and seriousness of the quality issues we uncovered accumulated.

The main categories of quality problems that exist within the original VRD visual relationship annotations are as follows:
\begin{enumerate}[(1)]
    \item stark variability in the types of objects sharing certain object class names;
    \item different object class names for objects clearly drawn from the same distribution and otherwise indistinguishable from one another;
    \item diverse semantics (usage) of single predicate names;
    \item outright errors in visual relationship construction;
    \item multiple near duplicate bounding boxes for the same object in a single image;
    \item multiple near and/or exact duplicate visual relationships annotated for an image;
    \item poor quality bounding boxes.
\end{enumerate}

\noindent
Examples of category (1) problems are:
\begin{itemize}
    \item object class \texttt{bear} pertains to real bears and teddy bears;
    \item object class \texttt{plate} pertains to dishware plates, vehicle license plates, baseball `home' plates and bases, plaques on walls, etc.;
    \item object class \texttt{glasses} pertains to eyeglasses, drinking glasses, miscellaneous things made of glass (e.g. table tops, glass barriers, etc.);
    \item object class \texttt{person} pertains to people and to stereo (audio) speakers.
\end{itemize}

\noindent
Examples of category (2) problems are:
\begin{itemize}
      \item object classes \texttt{plane} and \texttt{airplane} label objects that are indistinguishable;
      \item object classes \texttt{coat} and \texttt{jacket} do the same;
      \item object classes \texttt{road} and \texttt{street} do the same.
\end{itemize}

\noindent
Examples of category (3) problems are:
\begin{itemize}
  \item predicate \texttt{across} is used in the sense of `along side of', `is crossing the', `is across from', etc.;
  \item predicate \texttt{fly} is used in the sense of `is flying a', `is flown by', `is flying in`, `is flying above', etc.;
  \item predicate \texttt{walk} is used in the sense of `walk on', `walk next to', `walk towards each other', etc..
\end{itemize}

\noindent
Examples of category (4) problems are:
\begin{itemize}
  \item in possessive relationship patterns such as \texttt{(person, wear, X)} or \texttt{(person, hold, X)}, frequently the bounding box for \texttt{X} places it on/with a different person;
  \item in positional relationship patterns such as \texttt{(X, behind, Y)} or \texttt{(X, above, Y)}, frequently either \texttt{X} and \texttt{Y} need to be swapped, or the predicate changed to its inverse.  
\end{itemize}

\subsubsection{Impact of the VRD shortcomings.}

The VRD annotation quality problem categories just discussed have real consequences for the application task of visual relationship detection and will have (silently) negatively impacted all of the research that has been undertaken using the VRD dataset. 
Problem categories (1), (2), (5) and (7) hamper reliable object detection,
while categories (3), (4) and (6) have a negative impact on the detection of relationships.

The problem categories (1) and (3) also had real consequence for our objective of engineering a robust and credible OWL ontology to act as companion to the VRD dataset to enable our NeSy research using knowledge graphs. 
Together, problem categories (1) and (3) were serious enough that they effectively made the modeling of a credible ontology impossible.

We faced a tough choice between two awkward options. 
Option \textit{(i)} was to abandon the VRD dataset and continue searching for an image dataset with a matching ontology, whilst being uncertain as to our prospects for success at finding such a resource.
Option~\textit{(ii)} was to commit to investing the effort required to customise the VRD annotations so as to resolve the many issues uncovered by our deep analysis. 
At a minimum, this effort would need to be sufficient to resolve problem categories (1) and (3) to enable us to engineer a robust companion ontology.
This amount of effort might, however, also open the way to addressing the issues associated with the other problem categories as well with marginal incremental energy.
We selected option~\textit{(ii)}.
This decision led to the evolution of the resource that we call NeSy4VRD that is the subject of this paper.

\section{The NeSy4VRD dataset}

The NeSy4VRD image dataset is a substantially-improved version of the VRD image dataset~\cite{lu2016visual} introduced in 2016. 
The images of NeSy4VRD are the same as those of the original VRD dataset, but the NeSy4VRD visual relationship annotations for these images are a massively quality-improved version of the original VRD annotations.

\subsection{The images of the NeSy4VRD dataset}

The full VRD dataset stopped being publicly available around late 2021.
Given the popularity it has enjoyed in the past, it is very likely to be missed, despite its many serious (but, likely, largely unknown) shortcomings (discussed previously).
The original (problematic) VRD visual relationship annotations are still publicly available (through the website associated with \cite{lu2016visual}), but the VRD images themselves are no longer accessible from the URL specified in the documentation (readme.txt file) accompanying those annotations.

We recently contacted Dr. Ranjay Krishna, one of the principal originators of the VRD dataset and principal authors of \cite{lu2016visual}, about this matter.
Dr. Krishna has graciously granted us permission to host the VRD images ourselves so as to re-establish their public availability as part of NeSy4VRD.
Appropriate attribution for the VRD images is given as part of our NeSy4VRD dataset which is registered on Zenodo, at \url{https://doi.org/10.5281/zenodo.7916355}.

\subsection{The NeSy4VRD visual relationship annotations}

NeSy4VRD has comprehensively addressed and resolved the many serious shortcomings of the original VRD visual relationship annotations, outlined above in Section \ref{sec:vrd-problems}. 
As already mentioned, the prime motivation for resolving these shortcomings was to open the way for us to engineer a robust and credible OWL ontology as a well-aligned companion to the NeSy4VRD dataset.
This objective was achieved, and the NeSy4VRD VRD-World OWL ontology is the outcome of that effort.

To enable us to undertake this comprehensive customisation and quality-improvement exercise in a responsible fashion, we developed the NeSy4VRD protocol (for specifying visual relationship annotation customisations declaratively, in text files) and the NeSy4VRD workflow (for applying these customisations in a managed, automated and repeatable process).
These components, the NeSy4VRD protocol and workflow, are discussed shortly, in Section \ref{sec:extensibility}, as part of the explanation of the support that exists within NeSy4VRD for extensibility of the visual relationship annotations and, thereby, of the VRD-World ontology as well.

As will become clear in Section~\ref{sec:extensibility}, the NeSy4VRD protocol allows one to specify instructions to \textit{change} existing annotations, \textit{remove} unwanted annotations, and \textit{add} new visual relationship annotations for any VRD image.
The visual relationship annotation customisation instructions we specified using the NeSy4VRD protocol applied adjustments (changes, removals and additions) to the annotations of 1,715 of the 4,000 training images and to 828 of the 1,000 test images, for a total of 2,543 (just over half) of the 5,000 VRD images.

Table \ref{tab:anno-comparison} compares the NeSy4VRD visual relationship annotations with the original VRD annotations in terms of various salient statistics.
It shows the effect of the extensive annotation customisation exercise we undertook that can be conveyed quantitatively.
The table indicates that the NeSy4VRD annotations have more object classes, more predicates, more overall visual relationship annotations, and greater average annotations per image.
Most of the (arguably, more valuable) \textit{quality} improvements implemented as a result of our comprehensive annotation customisation exercise cannot be conveyed quantitatively in a table and remain implicit, but should not go unappreciated.

\begin{table}[t]
\caption{Quantitative comparison of the NeSy4VRD and VRD visual relationship annotations. `Visual relationship' has been abbreviated with the acronym VR.}
\label{tab:anno-comparison}
\begin{center}
\begin{tabular}{l|r|r}
\textbf{Metric} & \textbf{NeSy4VRD} & \textbf{VRD}  \\
\hline
Object classes & 109 & 100 \\
Predicates &  71 & 70 \\
\hline
Number of training set VR annotations & 29,333 & 30,355 \\
Number of test set VR annotations & 9,201 & 7,638 \\
Total number of VR annotations & 38,534 & 37,993 \\ 
\hline
Average VR annotations per training image & 7.8 & 7.6 \\ 
Average VR annotations per test image & 9.9 & 7.6 \\
\hline
Number of training images with duplicate VRs & 0 & 323  \\ 
Number of test images with duplicate VRs & 0 & 91  \\
\end{tabular}
\end{center}
\end{table}

\section{The NeSy4VRD OWL Ontology: VRD-World}

In this Section we describe the OWL ontology that we engineered to be a well-aligned companion of the NeSy4VRD dataset.
We call this ontology VRD-World because it directly describes the domain of the NeSy4VRD dataset as reflected in the 109 object classes and 71 predicates referenced in the NeSy4VRD visual relationship annotations of the VRD images.
The number and diversity of the object classes and predicates made it feasible to design the VRD-World ontology so as to have a reasonably rich class and property hierarchies which offer good opportunity for OWL reasoning capabilities to be meaningfully exercised.
Table \ref{tab:onto-metrics} gives a summary, quantitative view of the VRD-World ontology in terms of key metrics.

\begin{table}[t]
\caption{Summary metrics for the VRD-World ontology}
\label{tab:onto-metrics}
\begin{center}
\begin{tabular}{l|r}
\hline
\textbf{Summary Metric} & \textbf{Count} \\
\hline
Axiom & 815 \\
Logical axioms &  433 \\
Declaration axioms & 322 \\
Classes & 239 \\
Object properties & 74 \\
Data properties & 4 \\ 
Annotation properties & 8 \\
\hline 
\end{tabular}
\quad
\begin{tabular}{l|r}
\hline
\textbf{Class axioms} & \textbf{Count} \\
\hline
SubClassOf & 242 \\
EquivalentClasses & 21 \\
\hline
\textbf{Object property axioms} &  \\
\hline 
SubObjectPropertyOf & 46 \\ 
EquivalentObjectProperties & 7 \\
InverseObjectProperties & 6 \\ 
TransitiveObjectProperties & 17 \\ 
SymmetricObjectProperties & 8 \\
\hline
\end{tabular}
\end{center}
\end{table}

\begin{figure}[t!]
\includegraphics[width=\textwidth]{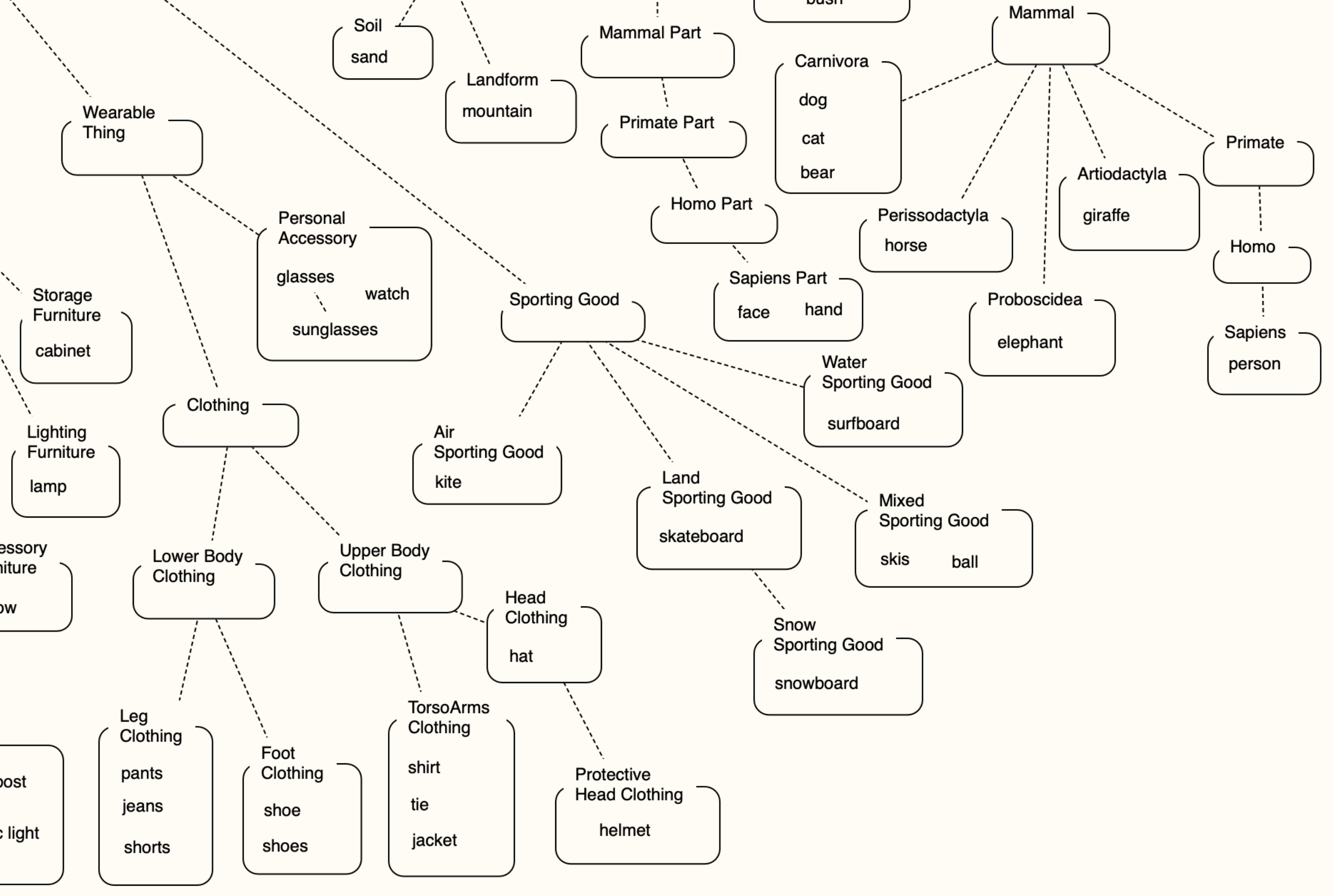}
\caption{A portion of the VRD-World class hierarchy.} 
\label{fig:class-hier}
\end{figure}

\medskip
\noindent
\textbf{Class hierarchy}. Figure \ref{fig:class-hier} gives a pictorial representation of a small portion of the VRD-World class hierarchy.
Names situated at the top edge of a bubble (like \texttt{Personal Accessory}) represent class names that we introduced as part of organising the \linebreak NeSy4VRD object classes into a reasonable hierarchy.
Names situated inside bubbles (like \texttt{sunglasses} and \texttt{watch}) are NeSy4VRD object classes, and the fact of being `inside' represents an implicit \texttt{rdfs:subClassOf} relationship.
All dotted lines represent explicit \texttt{rdfs:subClassOf} relationships.

\begin{figure}[t]
\includegraphics[width=0.99\textwidth]{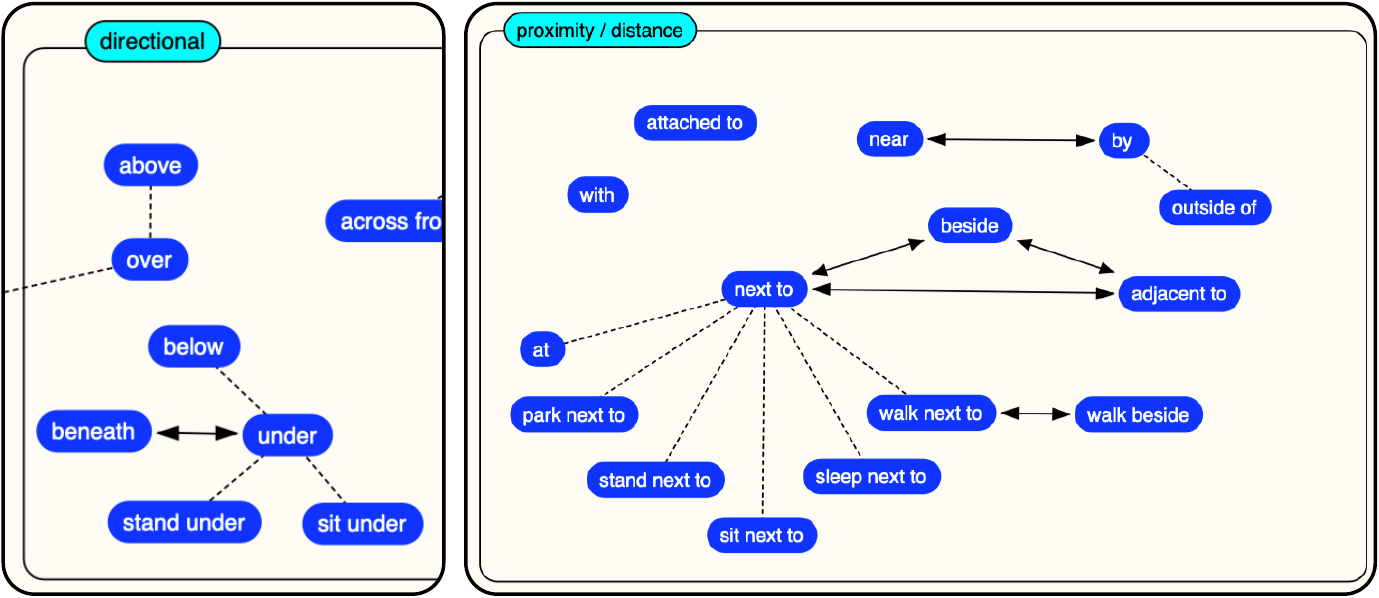}
\caption{Two portions of the VRD-World object property hierarchy.} 
\label{fig:prop-hier}
\end{figure}

\medskip
\noindent
\textbf{Property hierarchy}. Figure \ref{fig:prop-hier} gives a pictorial representation of two small portions of the VRD-World object property hierarchy.
All names surrounded in dark blue are NeSy4VRD predicate names that have directly corresponding object properties in the VRD-World ontology.
Dotted lines represent \texttt{rdfs:subPropertyOf} relationships and dark, double-headed arrows represent \texttt{owl:equivalentClass} relationships.
Several \texttt{owl:inverseOf} relationships exist but are not represented in the diagram to minimise clutter.
For example, properties \texttt{above} and \texttt{below} are inverses, as are \texttt{over} and \texttt{under}.

\begin{figure}[t]
\centering
\includegraphics[width=0.9\textwidth]{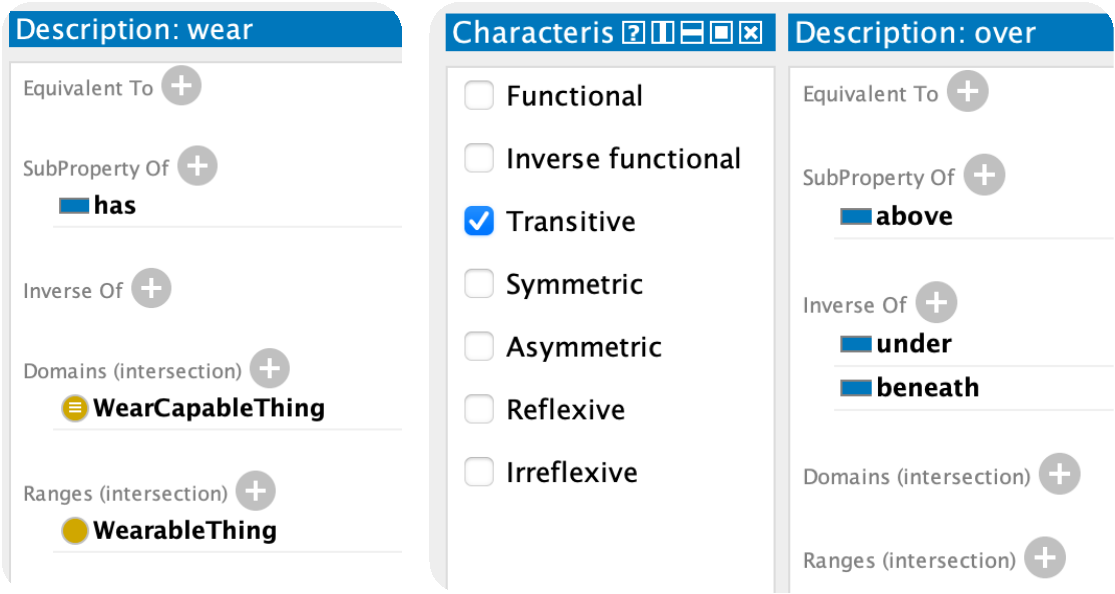}
\caption{Two object properties of VRD-World.} 
\label{fig:props}
\end{figure}

\medskip
\noindent
\textbf{Property descriptions and characteristics}. Figure \ref{fig:props} shows two object properties of VRD-World as they appear in the Prot\'eg\'e ontology editor. Most of the characteristics they possess are reflected in the summary metrics displayed in Table \ref{tab:onto-metrics}. These examples help to justify the claim made earlier that VRD-World is a rich ontology capable of exercising OWL reasoning capabilities meaningfully.

%

\section{Infrastructure to load and extract annotations}

The VRD-World ontology describes more of the NeSy4VRD domain than just the object classes and predicates referenced in the annotations: it describes conceptual infrastructure sufficient to capture the NeSy4VRD visual relationship annotations in their entirety, including representations of the images (labelled with their unique filenames) and the bounding box specifications of the objects.
As in the annotations, in the ontology each object (with its bounding box) is anchored to a unique image (via an inverse-functional \texttt{:hasObject} object property).
So, in a knowledge graph hosting the VRD-World ontology, each data instance representing a VRD image is effectively the root of a distinct subgraph and can be leveraged in that capacity.

This feature of VRD-World permits users to do things such as \textit{(i)} load the NeSy4VRD visual relationship annotations into a knowledge graph (in their entirety, or for a single image), \textit{(ii)} materialise the knowledge graph (by appropriate means) to infer new visual relationships, and \textit{(iii)} extract 
an augmented set of visual relationship annotations, in their native format. 
NeSy4VRD provides sample code for performing these load and extract tasks using the well-known Python package RDFLib.
The rationale for this is to further lower the barrier to entry to NeSy research using OWL-based knowledge graphs for researchers who may be tempted to pursue it.

\section{NeSy4VRD support for extensibility}
\label{sec:extensibility}

In this Section we describe the components of NeSy4VRD that together provide comprehensive support for the extensibility of the NeSy4VRD visual relationship annotations and, thereby, of the VRD-World ontology as well.
The NeSy4VRD visual relationship annotations and companion VRD-World OWL ontology can of course be used \textit{as is}.
This is what we do in our NeSy research using OWL-based knowledge graphs.
But these two aspects of NeSy4VRD have, from the outset, been regarded by us as \textit{defaults}.
We anticipate that, oftentimes, NeSy4VRD may be perceived by researchers as being close to what they are looking for in a research resource, but that their particular research needs and interests are such that the NeSy4VRD visual relationship annotations and companion OWL ontology, VRD-World, require some amount of tailoring in order that they suit the particularities of their research needs sufficiently well.
The rationale for NeSy4VRD providing the extensibility-supporting components that we describe here is to lower the barrier to performing such tailoring, should it be required.

\subsection{Comprehensive code for dataset analysis}

A prerequisite for customising the NeSy4VRD visual relationship annotations in a systematic way is being able to thoroughly analyse them, and in conjunction with their associated VRD images.
NeSy4VRD provides comprehensive Python code for doing just this.
This is the same code we developed for undertaking our own deep analysis of the VRD images and original VRD visual relationship annotations, and which led to the discovery of the many issues and shortcomings latent in those annotations, as detailed in Section~\ref{sec:vrd-problems}. 

We list a small sampling of the basic functionality available in order to give a flavour of the analytic facilities NeSy4VRD provides:
\begin{itemize}
    \item display an image, showing the objects annotated for: \textit{(i)} one particular visual relationship (VR), \textit{(ii)} a particular subset of VRs, \textit{(iii)} all VRs;
    \item print all VRs for an image: \textit{(i)} in readable \texttt{(s, p, o)} form, \textit{(ii)} in raw form;
    \item find all images having VRs for: \textit{(i)} object class \texttt{X} or object classes \texttt{[X,Y,...]}, \textit{(ii)} a particular predicate, \textit{(iii)} a particular VR pattern, such as \texttt{(s, p, X)}, \texttt{(X, p, o)}, \texttt{(s, X, o)}, \texttt{(s, p, o)}, and list the distinct values for X;
    \item find all images with a target number of VRs
    \item distributional analyses: \textit{(i)} number of VRs per image, \textit{(ii)} number of distinct object classes per image, \textit{(iii)} number of distinct predicates per image, etc.;
    \item quality verification analyses: find images with \textit{(i)} duplicate VRs, \textit{(ii)} degenerate bounding boxes, \textit{(iii)} bounding boxes assigned multiple object classes, etc.
\end{itemize}

\subsection{The NeSy4VRD protocol}

To enable large volumes of visual relationship annotation customisations to be applied safely and using an automated, repeatable mechanism, NeSy4VRD uses a custom protocol so that they can be specified declaratively, in text files. We call this the NeSy4VRD protocol. Listing \ref{lst:protocol} shows some representative customisation instructions associated to five images and the different instruction types defined for the protocol.

\begin{lstlisting}[caption={Example NeSy4VRD protocol annotation customisation instructions.}, label={lst:protocol}, basicstyle=\ttfamily\small]
imname; 3223670633_7d3d72dfe8_b.jpg
cvrsoc; 4; (`person', `on', `shelf'); speaker
cvrsbb; 4; (`speaker', `on', `shelf'); [161,234,231,270]

imname; 8934043045_251b42d19a_b.jpg
cvrooc; 7; (`bus', `beside', `car'); truck
cvrobb; 7; (`bus', `beside', `truck'); [334,557,99,403]

imname; 1426904233_ee344879b6_b.jpg
cvrsoc; 5; (`bear', `sit on', `basket'); teddy bear
cvrpxx; 5; (`teddy bear', `sit on', `basket'); in

imname; 4929276486_ca06aedbb9_b.jpg
rvrxxx; 4; (`person', `wear', `jacket');
avrxxx; boat; [477,594,319,746]; has; dog; [478,529,587,618]
avrxxx; boat; [477,594,319,746]; carry; dog; [478,529,587,618]

imname; 7171463996_900cb4ce33_b.jpg; rimxxx
\end{lstlisting}

For the first image in the listing, an improvement to the visual relationship annotation at index 4 (in the image's list of annotations) is being specified.  The \texttt{cvrsoc} instruction declares an intention to change the subject's object class (\texttt{soc}) from \texttt{person} to \texttt{speaker}, (as in `stereo speaker'). The \texttt{cvrsbb} instruction declares an intention to change the subject's bounding box (\texttt{sbb}) to a more accurate localisation of the object.

For the second image, similar types of changes are being specified, but this time for the visual relationship at index 7 and in relation to the object's object class and bounding box (\texttt{cvrooc} and \texttt{cvrobb}).

For the third image, an improvement to the visual relationship at index 5 is being specified. After a \texttt{cvrsoc} instruction specifying that the subject's object class be changed from \texttt{bear} to \texttt{teddy bear}, a \texttt{cvrpxx} instruction declares an intention to change the predicate (\texttt{pxx}) from \texttt{sit on} to \texttt{in}.

For the fourth image, the \texttt{rvrxxx} instruction declares an intention to remove the visual relationship at index 4 (because, say, it was found to be too badly broken or to be a near or exact duplicate of another annotation for the same image). The two \texttt{avrxxx} instructions declare intentions to \textit{add} two \textit{new} visual relationships to the set of annotations for the image.

Finally, the \texttt{rimxxx} instruction following the name of the fifth image in the listing declares an intention to have the entry for that image removed from the annotations dictionary (because the image and its annotations were found to be highly problematic and not recoverable).

A dedicated Python script processes an annotation customisation instruction text file sequentially, interprets the protocol, validates the construction of the annotation customisation instructions, and executes them.  If an error is detected, execution is aborted and the cause of the problem and the line number of the offending instruction are reported to the user.
We used GIMP (\url{gimp.org}), the free and open source image editor, to determine all bounding boxes specified in our annotation customisation instructions. But users of NeSy4VRD can of course select whichever tool they prefer for this subtask.

\subsection{The NeSy4VRD workflow}

The NeSy4VRD workflow is a set of Python modules and scripts that implement a configurable, multi-step sequential process for applying one's planned and pre-specified visual relationship annotation customisations in a configurable, automated and repeatable manner. 
Each sequential step of the workflow relates to discrete category of annotation customisation and is performed by a dedicated Python script designed for that task.
The data governing the precise annotation customisations applied by each sequential step in a particular execution of the workflow are defined in a Python configuration module using step-specific, predetermined variable names and formats.
Each step (script) of the workflow imports this configuration module to  access the variables it needs in order to execute properly.
There are two such configuration modules: one for managing the running of the workflow to apply customisations to the annotations of the VRD training images, and one for the test images.
The precise set of workflow steps needed when running the workflow against the training annotations and test annotations need not be identical.
The responsibilities of the successive steps of the NeSy4VRD worflow that we used for applying our NeSy4VRD customisations to the original VRD annotations are as follows:
\begin{enumerate}
    \item change existing and/or add new object class or predicate names to the respective master lists that define these names;
    \item apply the annotation customisation instructions that have been specified in a particular text file using the NeSy4VRD protocol;
    \item for a specified set of images, change all instances of object class X to object class Y; (more than one set of images can be processed in this way);
    \item merge all instances of object class X into object class Y; merge all instances of predicate X into predicate Y; (multiple such pairs can be processed);
    \item remove all instances of specified VR \textit{types}, globally;
    \item remove all image entries from the annotations dictionary with zero VRs;
    \item apply the annotation customisation instructions that have been specified in a particular text file using the NeSy4VRD protocol;
    \item apply the annotation customisation instructions that have been specified in a particular text file using the NeSy4VRD protocol;
    \item change VR \textit{type} X to \textit{type} Y, globally; (multiple such pairs can be processed; restrictive conditions apply);
    \item find images with duplicate VRs and remove the duplicates, globally;
    \item apply the annotation customisation instructions that have been specified in a particular text file using the NeSy4VRD protocol.
\end{enumerate}

\section{NeSy4VRD beneficiaries and use cases}

In this Section we briefly highlight a few intended and potential NeSy4VRD beneficiaries and use cases.
NeSy4VRD has a real prospect of being attractive to diverse user groups.
The primary users of NeSy4VRD are expected to be:
\begin{itemize}
    \item computer vision researchers interested in using NeSy4VRD as a quality-improved version of the (now partly unavailable) VRD dataset, but for deep learning alone, ignoring the companion VRD-World OWL ontology;
    \item NeSy researchers not using Semantic Web technologies but who nonetheless find NeSy4VRD attractive for NeSy research purposes, for reasons such as those outlined in Section \ref{sec:vrd-attractive} (small size, and zero/few-shot learning);
    \item NeSy researchers using Semantic Web technologies such as OWL ontologies and OWL-based knowledge graphs who are looking for a dataset with a well-aligned, companion OWL ontology to enable them to pursue their research vision.
\end{itemize}

The paramount intended use case for NeSy4VRD is vision-based NeSy research that relies on leveraging OWL ontologies in some way. 
The manner in which the NeSy4VRD dataset and its companion VRD-World OWL ontology might be leveraged in this context is limited only by the imaginations of NeSy researchers.
The literature of NeSy research using Semantic Web technologies has examples of OWL ontologies being leveraged from a structural perspective (irrespective of their ability to drive deductive reasoning) to enhance image classification.
One instance of this is \cite{DBLP:conf/www/GengC0PYYJC21}, which uses embeddings of ontologies to enhance zero-shot learning in image classification.
Our own research has a particular focus on exploring how OWL and Datalog knowledge graph reasoning, driven and governed by our VRD-World ontology, can be leveraged to enhance deep learning for visual relationship detection \cite{herron2022}.
NeSy4VRD and its VRD-World ontology are also well positioned to participate in the emerging NeSy subfield of symbolic knowledge injection/infusion \cite{DBLP:journals/internet/KhanBC22,PsykiExtraamas2022} 

Another potential category of use case for NeSy4VRD is as a standard or benchmark resource for vision-based research, especially for NeSy research using OWL ontologies and knowledge graphs.
The popularity of the VRD dataset suggests it became something of a standard or benchmark dataset for the application tasks of visual relationship detection and scene graph generation.
Given that NeSy4VRD is a quality-improved version of the (no longer fully available) VRD dataset, it is entirely plausible that NeSy4VRD inherits that role.
And given the scarcity of research resources like NeSy4VRD that target the specialised needs of NeSy researchers wishing to use OWL ontologies and knowledge graphs, NeSy4VRD has strong potential for becoming a standard or benchmark resource for that user group in particular.
Finally, given its comprehensive support for extensibility, it is conceivable that the default NeSy4VRD visual relationship annotations and companion VRD-World OWL ontology give rise to an ever-growing \textit{family} of unique but strongly related standard dataset resources.

\section{Concluding remarks}

NeSy4VRD is a multifaceted, multipurpose research resource.
It has an image dataset for visual relationship detection accompanied by a well-aligned OWL ontology describing the dataset domain. 
It provides comprehensive support for extensibility of the image annotations and, thereby, of the OWL ontology.
And it provides sample code for loading the annotations to/from a knowledge graph.

NeSy4VRD makes the VRD images available again and in conjunction with a massively quality-improved version of the VRD visual relationship annotations.
Like the VRD dataset, NeSy4VRD has characteristics (small size, and plentiful zero/few-shot learning data conditions) that are likely to be especially attractive for NeSy researchers.
It specifically addresses the special needs of NeSy researchers wishing to use OWL ontologies and knowledge graphs as symbolic background knowledge and reasoning components for whom appropriate research resources are particularly scarce.
NeSy4VRD has strong prospects for becoming a standard resource for diverse user groups and we are pleased to contribute it to the computer vision, NeSy and Semantic Web communities.
In so doing, we particularly hope to foster more NeSy research using OWL-based knowledge graphs.

\subsection*{Acknowledgements}

We would like to thank Dr. Ranjay Krishna for granting us permission to re-establish the public availability of the VRD images as part of NeSy4VRD.
This research was partially funded by the SIRIUS Centre for Scalable Data Access (Research Council of Norway, project 237889).

%
%
%
\bibliographystyle{splncs04}

\bibliography{main}

\end{document}